\documentclass[conference]{IEEEtran}
\IEEEoverridecommandlockouts
\usepackage{cite}
\usepackage{amsmath,amssymb,amsfonts}
\usepackage{algorithmic}
\usepackage{graphicx}
\usepackage{textcomp}
\usepackage{xcolor}
\usepackage{float}
\usepackage{multirow}
\usepackage{array}
\usepackage{float}
\usepackage{placeins}

\def\BibTeX{{\rm B\kern-.05em{\sc i\kern-.025em b}\kern-.08em
    T\kern-.1667em\lower.7ex\hbox{E}\kern-.125emX}}
\begin{document}

\title{Effects of Different Prompts on the Quality of GPT-4 Responses to Dementia Care Questions
}


\author{\IEEEauthorblockN{Zhuochun Li\IEEEauthorrefmark{1},
Bo Xie\IEEEauthorrefmark{2}, Robin Hilsabeck\IEEEauthorrefmark{3}, Alyssa Aguirre\IEEEauthorrefmark{4}, Ning Zou\IEEEauthorrefmark{1}, Zhimeng Luo\IEEEauthorrefmark{1} and Daqing He\IEEEauthorrefmark{1}}
\IEEEauthorblockA{\IEEEauthorrefmark{1}School of Computing and Information, University of Pittsburgh, Pittsburgh, USA}
\IEEEauthorblockA{\IEEEauthorrefmark{2}School of Nursing and School of Information, University of Texas at Austin, Austin, USA}
\IEEEauthorblockA{\IEEEauthorrefmark{3}University of Texas Health Sciences Center at San Antonio, San Antonio, TX, USA}
\IEEEauthorblockA{\IEEEauthorrefmark{4}Dell Medical School, University of Texas at Austin, Austin, USA}
\IEEEauthorblockA{Email: zhl163@pitt.edu,
boxie@utexas.edu,
hilsabeck@uthscsa.edu, \\
alyssa.aguirre@austin.utexas.edu,
niz19@pitt.edu,
zhl123@pitt.edu,
dah44@pitt.edu}
}

\maketitle

\begin{abstract}

Evidence suggests that different prompts lead large language models (LLMs) to generate responses with varying quality. Yet, little is known about prompts' effects on response quality in healthcare domains. In this exploratory study, we address this gap, focusing on a specific healthcare domain: dementia caregiving. We first developed an innovative prompt template with three components: (1) system prompts (SPs) featuring 4 different roles; (2) an initialization prompt; and (3) task prompts (TPs) specifying different levels of details, totaling 12 prompt combinations. Next, we selected 3 social media posts containing complicated, real-world questions about dementia caregivers’ challenges in 3 areas: memory loss and confusion, aggression, and driving. We then entered these posts into GPT-4, with our 12 prompts, to generate 12 responses per post, totaling 36 responses. We compared the word count of the 36 responses to explore potential differences in response length. Two experienced dementia care clinicians on our team assessed the response quality using a rating scale with 5 quality indicators: factual, interpretation, application, synthesis, and comprehensiveness (scoring range: 0-5; higher scores indicate higher quality). Both clinicians rated the responses from 3 to 5, with 75\% agreement. Consensus was reached through discussion. Overall, 44\% of responses (16/36) were rated as 5; another 44\% (16/36), as 4; the remaining 4 (11\%), as 3. We found no interaction effect of system and task prompts or main effect of system prompts on response length. Task prompts had a statistically significant effect on response length: \textit{F}(2,24) = 82.784, \textit{p} \textless .001. Post hoc analysis showed that the significant difference in responses was due to TP3, which led to significantly longer responses. There was no interaction or main effect of system and task prompts on response quality. Our clinicians' qualitative feedback provided further insight: (1) system prompts with the different professional roles (neuropsychologist and social worker) did not lead to noticeable differences in response content (that is, there were \textit{no} neuropsychology- and social work-\textit{versions} of GPT-4 responses); and (2) TP3, while producing longer responses \textit{statistically}, might not necessarily have produced higher quality responses \textit{clinically}: at times the details contained in the lengthy responses seem \textit{unnecessary} from a clinical perspective. We discuss study limitations and future research directions. 
\end{abstract}

\begin{IEEEkeywords}
large language models, prompt engineering, dementia, informal caregiving, social media
\end{IEEEkeywords}

\section{Introduction}

Dementia has become a major public health challenge \cite{shu2021use}. Persons with dementia require extensive care, often provided by family members who frequently report experiencing challenges and stress~\cite{alz2022,reid2021psychological}. Artificial intelligence (AI) may support caregivers; but research is needed to understand the ways in which AI might be effective ~\cite{ji2022automatic}.
 
Large language models (LLMs) such as ChatGPT~\cite{wu2023brief} and GPT-4~\cite{achiam2023gpt} can comprehend and interpret users' needs, and generate outputs that meet their requirements
~\cite{brown2020language, moor2023foundation}. Compared to traditional few-shot learning methods in clinical NLP~\cite{li2023siakey},  LLM can generate high-quality responses to dementia caregivers' questions to help overcome the challenges caregivers experience~\cite{alyssa2023assessing}.


A prompt is a natural language command that instructs a large language model (LLM) to complete a user-specified task~\cite{wang2023prompt}. Prompt engineering refers to the action of designing prompts to instruct LLMs to generate responses highly pertinent to users' inquiries in specialized domains~\cite{henrickson2023prompting, white2023prompt}. Preliminary evidence shows that different prompts influence the quality of LLMs' outputs~\cite{wang2023prompt}. Yet little is known about the effects of different prompts on the quality of LLM-generated responses in healthcare domains. 

\begin{figure}[htb] 
\centering 
\includegraphics[width=0.35\textwidth,height=0.5\textwidth]{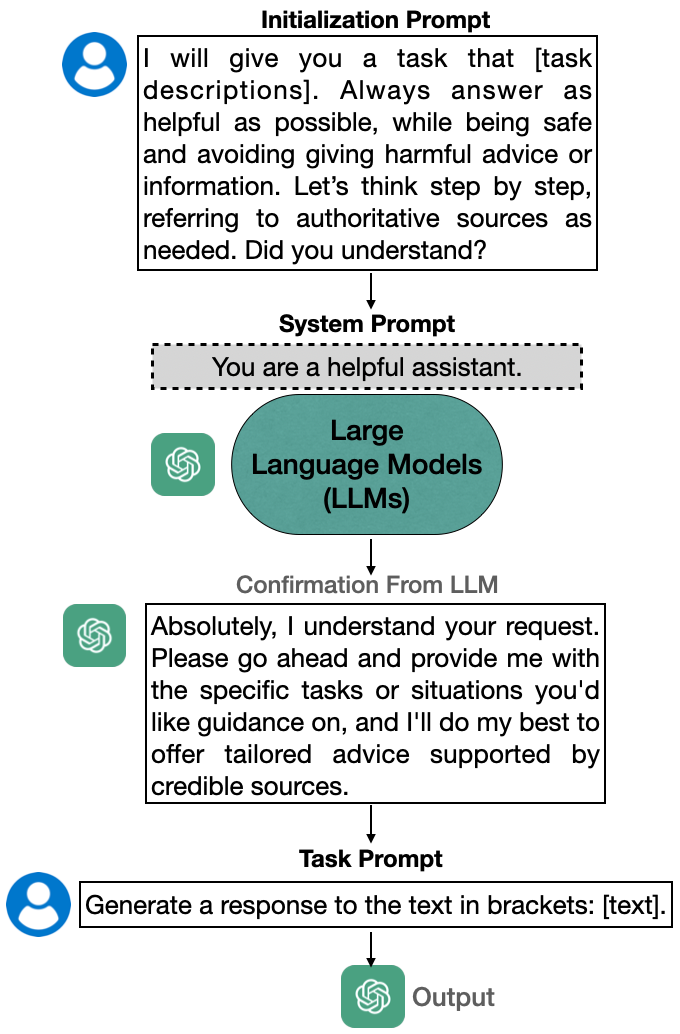} 
\caption{Overview of our Prompt Template} 
\label{fig1} 
\vspace{-10pt}
\end{figure}

In this exploratory study, we focus on dementia care as an example to illustrate whether and how prompt engineering might affect healthcare practice. We use GPT-4 as the LLM and examine the combinations of three types of prompts: system prompts, initialization prompt, and task prompts (Fig. ~\ref{fig1}). System prompts, common in the literature, refer to default prompts set to the LLM before starting a conversation~\cite{zheng2023helpful}. In a previous study~\cite{zheng2023helpful}, LLMs performed differently when their role was uniquely defined in the system prompt. Upon review of ChatGPT's prompt engineering website~\cite{chatgpt_official}, we also found a similar indication: ``Tactic: Ask the model to adopt a persona.`` We define the \textit{initialization prompt} as instructions that prepare the LLM for tasks ahead (the LLM's response to the initialization prompt helps confirm that it understands and is ready for the task). Task prompts are specific instructions in which users instruct LLMs what specific tasks to perform~\cite{wang2023prompt} GPT-4 generated responses are evaluated by experienced dementia care clinicians on our team. Tactics on the ChatGPT's prompt engineering website~\cite{chatgpt_official} such as ``Use delimiters to clearly indicate distinct parts of the input,`` ``Specify the steps required to complete a task,`` and ``Specify the desired length of the output`` informed the development of our task prompts. Findings of this exploratory study have implications for future research on prompt engineering in healthcare.

\section{Related Work}

With recent developments in computational capabilities and available datasets, LLMs trained on general domain data have demonstrated substantially improved performance, eliminating the need for domain-specific pretraining. Kung et al.~\cite{kung2023performance} evaluated ChatGPT-3.5 responses to questions from the United States Medical Licensing Exam (USMLE) and achieved results at or near the passing threshold without any specialized training. Nori et al.~\cite{nori2023capabilities} demonstrated that GPT-4 surpassed the USMLE passing score by over 20 points using straightforward 5-shot prompting. 

Prompt-based techniques have been used for a wide range of real world tasks~\cite{wang2023prompt}. 
In healthcare domains, DeID-GPT~\cite{liu2023deid} has
leveraged high-quality prompts to safeguard privacy and summarize critical information within medical data, achieving outcomes better than those obtained with several baseline methods. HealthPrompt~\cite{sivarajkumar2022healthprompt}, a prompt-based clinical natural language processing framework exploring various prompt templates, has been shown to enhance 
performance in clinical text classification tasks without necessitating additional training data.

LLMs' performance for specific tasks can be notably influenced by the prompt used~\cite{nori2023can}. 
Chain of Thought (CoT) is a prompting approach that incorporates intermediate reasoning steps before presenting an answer~\cite{wei2022chain}. By decomposing intricate problems into a sequence of smaller steps, CoT assists in building a foundation model to generate precise answers~\cite{singhal2022large}. Different social roles specified in prompts were also found to have impacts to LLMs' responses~\cite{zheng2023helpful}. Through studying 162 social roles such as ``mother``, ``layer``, and ``police,`` on a set of language understanding tasks specified by dataset MMLU (Massive Multitask Language Understanding), Zhen et al.~\cite{zheng2023helpful} showed that certain roles improved LLMs' prediction accuracy on these language understanding tasks.

\section{Study Design 
}
This study involved 3 steps: (1) selecting dementia care posts to be used for queries with LLMs; (2) designing the prompts; and (3) assessing the responses’ quality.

\subsection{Selecting Dementia care posts}
We first selected dementia care posts that contained caregivers' questions (that were then used, with different prompts, for GPT-4 to generate responses to the questions). The posts were collected from a social media platform, Reddit. Reddit was chosen due to its prevalence among individuals sharing dementia-related challenges~\cite{zou2023mapping, tang2023online}. The abundance of relevant posts, facilitated by the Reddit API, has enabled earlier research to identify and examine caregivers' health information wants (HIWs), i.e., the types and amounts of healthcare information that caregivers wish to have to manage the care for their family members living with Alzheimer's disease and related dementias~\cite{wang2021characterizing}. 

For this study, we selected three posts, each one representing a prevalent type of HIW that caregivers wish to have in their daily care for persons with dementia: memory loss and confusion, aggression, and driving. As one example, below is the Reddit post we selected for the HIW category of memory loss and confusion:

\textit{My grandmother has lived in the same house for 57 years, and now she is convinced that we've moved to a house ``down the shore`` She is in the late stages of Alzheimer’s, and every night she asks me ``have you ever been here before?`` and I say of course and explain to her that we are still in the town we have lived in for over half a century, which just confuses her and then she will ask again in a minute or two. My question is what can I do to make her realize that she is home?}

\subsection{Designing the Prompts}
We included 3 types of prompts: system prompts (4 roles), an initialization prompt, and task prompts (3 tasks).

\subsubsection{System Prompts}
We developed 4 roles in the system prompt. The first is the default role in the system prompt for GPT-4: \textit{``You are a helpful assistant``.} We designed the other three roles to focus on our specific healthcare domain, dementia care: \textit{``an experienced clinician specialized in dementia care``,} \textit{``a board-certified clinical neuropsychologist``,} and \textit{``a licensed clinical social worker``.} These three roles represent common types of clinicians who may engage in consultations with persons living with dementia and their family caregivers (and reflect the expertise we have on our team). We hypothesized that these roles might trigger GPT-4 to shift its setting from a general area to healthcare domains, which might subsequently improve its performance on our tasks. We used GPT-4 API from openAI to issue system prompts.

\subsubsection{Initialization Prompt}
When constructing the \textit{initialization prompt}, we instructed the LLM clearly and specifically on what we view as high quality responses: 
\textit{``Always answer as helpful as possible, while being safe and avoiding giving harmful advice or information. Let’s think step by step, referring to authoritative sources as needed.``} Then by recognizing that our prompts are long, we designed in the initialization prompt to elicit a confirmation from the LLM: 
\textit{``Do you understand?``} 
If the LLM's confirmation indicates that it does not understand the task, we will continue with more instructions to help it. Thus the complete initialization prompt is \textit{``I will give you tasks that [specific task descriptions]. Always answer as helpful as possible, while being safe and avoiding giving harmful advice or information. Let’s think step by step, referring to authoritative sources as needed. Do you understand?``}

\subsubsection{Task Prompts}
Task prompt is the part of the prompt template we designed to include task specific instructions. For example, our task is related to generate care tips for answering the question raised by a caregiver.  Through working with our clinicians, we developed an answer structure with three important aspects of tips: educational information, tangible actions, and referral suggestions. Consequently, one task prompt (TP2) gives clear instruction to the LLM to generate the responses by following the answer structure.  
Furthermore, we another task prompt (TP3) to ask the LLM to generate responses without length limit: ``Do not limit the length of the response to any word limit`` to instruct GPT-4 to provide more details in its responses. Table~\ref{table1} shows the task prompts.

\begin{table}[hbt]
\caption{\label{table1} Examples of our task prompts. ``+`` is defined as adding words to the previous prompt.}
\centering
\begin{tabular}{p{2cm}|p{6cm}}
\hline \textbf{Task Prompt 1} & Generate a response to the text in brackets: [post]. \\ \hline
\textbf{Task Prompt 2} & + In generating your response, follow the following 3-component structure: first, educational information; second, tangible actions the caregiver can take; and finally, referrals that suggest helpful resources for the caregiver.\\ \hline
\textbf{Task Prompt 3} & + Try to generate the response as long as possible with as much detail as possible to cover all relevant aspects with all available specifics. Do not limit the length of the response to any word limit.\\ \hline
\end{tabular}
\vspace{-10pt}
\end{table}

\subsection{Assessing Response quality}
Two clinicians, each with 15+ years of experience with persons living with dementia and their family caregivers, independently evaluated the responses using a 5-point rating scale~\cite{alyssa2023assessing} that was previously developed to evaluate the quality of ChatGPT responses to dementia care posts. This scale assessed LLM-generated responses based on 5 quality indicators: \textit{factual; interpretation; application; synthesis; and comprehensiveness}. Each response was assessed as 0 (absent) or 1 (present) on each quality indicator (scoring range: 0-5; higher scores indicate higher quality). Differences between the two clinicians' ratings were resolved through discussion.

In addition to these quantitative evaluations, we conducted an inductive thematic analysis~\cite{braun2006using} of the responses to identify patterns. This effort was led by one of the coauthors (BX), an expert in qualitative data analysis methods including thematic analysis. The identified initial patterns were discussed with the clinicians, who provided feedback to help verify and refine the patterns. This effort focused on identifying components that should be included in the prompts in the future to improve the LLM's performance.

\section{Results}
We compared the word counts in the responses (Table~\ref{table2}). Two-way analysis of variance (ANOVA) found no interaction effect between system and task prompts [\textit{F}(6,24) = 0.673; \textit{p} = 0.673] or main effect of system prompts [\textit{F}(3,24) = 1.143; \textit{p} = 0.352] on response length. Task prompts had a statistically significant main effect on response length: \textit{F}(2,24) = 82.784, \textit{p} \textless 0.001. Post hoc analysis showed that the significant difference in responses was due to TP3, which led to the longest responses.

\begin{table}[hbt]
\centering
\caption{\label{table2} word counts in GPT-4 responses to different prompts.}
\scalebox{0.9}{
\begin{tabular}{|p{0.8cm}|p{0.5cm}|p{0.45cm}|p{0.65cm}|p{0.5cm}|p{0.45cm}|p{0.65cm}|p{0.5cm}|p{0.45cm}|p{0.65cm}|} \hline 
&  \multicolumn{3}{|c|}{Task Prompt\_1}&  \multicolumn{3}{|c|}{Task Prompt\_2}&  \multicolumn{3}{|c|}{Task Prompt\_3}\\ \hline 
&  min-max&  mean&  median&  min-max&  mean&  median&  min-max&  mean& median\\ \hline 
System Prompt1&  230-328&  277.3&  274&  263-379&  314.7&  302&  525-611&  567.3& 566\\ \hline 
System Prompt2&  292-323&  308.7&  311&  325-365&  341.7&  335&  449-578&  521.0& 536\\ \hline 
System Prompt3& 324-377& 353.7& 360& 298-393& 352.0& 365& 527-620& 567.0&554\\ \hline 
System Prompt4& 210-345& 296.7& 335& 336-367& 352.3& 354& 458-595& 527.3&529\\ \hline
\end{tabular}}
\end{table}

\begin{figure*}[hbt]
    \centering
     \includegraphics[width=1\textwidth,height=0.3\textwidth]{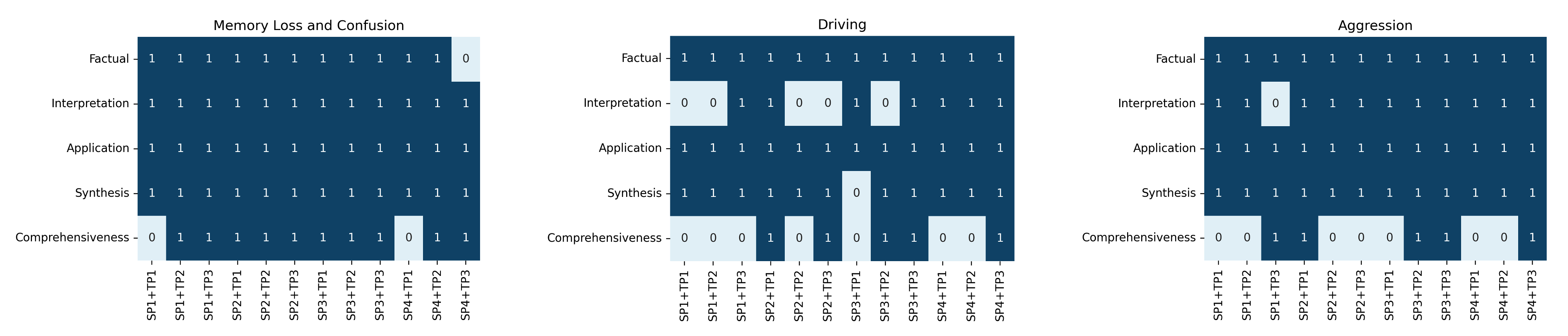}
     \caption{Clinicians' evaluation results.}
    \label{fig2}
\end{figure*}

Clinicians’ ratings of the responses ranged from 3 to 5; no response was rated lower than 3. They didn't identify any fabrication or hallucination in GPT-4 generated responses. The clinicians agreed on 27 of 36 response ratings (75\% agreement). Consensus was reached through discussion. After consensus, 44\% (16/36) of responses were rated as 5; another 44\% (16/36) were rated as 4; the remaining 4 (11\%) were rated as 3. These ratings are illustrated in Figure~\ref{fig2}. Two-way ANOVA found no interaction or main effect of the system and task prompts on response quality.

In addition to these quantitative findings, our inductive thematic analysis also identified patterns in GPT-4 responses. 
\begin{itemize}
    \item TP2 asked GPT-4 to generate responses following a specified structure containing 3 specific components (i.e., educational information, tangible actions, and referrals) in a specified order. We noted that all responses contained all 3 components, with or without TP2. However, with TP2, responses were explicitly structured, enhancing the responses’ structural consistency (and likely readability). 
    \item TP3 instructed GPT-4 to generate as much detailed information as possible. Responses to TP 3 were indeed longer than those to TP1 and TP2 across the posts, statistically. However, our clinicians noted that some of the educational information in the longer responses might be unnecessary. For example, one response to the post about memory loss and confusion recommended ``a full cognitive assessment,`` which the clinicians felt was unnecessary because the patient was already diagnosed with dementia. (That response was prompted by the combination of SP1 + TP3. It was the only case in which ``a full cognitive assessment`` was recommended.) In another case, the phrase ``deterioration of pragmatic communication competence`` was introduced as part of the educational information for the prompt combination of SP4 + TP3. Yet the clinicians felt that the use of such a phrase was ``off the mark`` and unnecessary; it did not address the caregiver’s question about how to care for a family member with memory loss and confusion. One of the clinician, a social worker, noted that she had \textit{never} used that phrase in all her consultation sessions with dementia patients and caregivers. In these cases, TP3 was the common prompt used.
    \item Recommendations in GPT-4’s responses to the post on driving were repetitive of what the caregivers had already tried. For example, GPT-4 recommended following up with a doctor or the Department of Motor Vehicles, even though the post had stated explicitly that such a strategy had already been tried (and had failed). Repetitive recommendations appeared in responses to multiple combinations of prompts.
    \item The clinicians did not identify noticeable differences in the responses to system prompts specifying the role of ``neuropsychologist`` (SP3) or ``social worker`` (SP4): there were \textit{no} neuropsychology- and social work-\textit{versions} of responses.
\end{itemize}

\section{Discussions}
In this exploratory study, we examined the effects of different combinations of prompts on the quality of GPT-4 responses to dementia caregivers’ questions. Overall, GPT-4 responses were of high quality, with the majority rated 5 or 4 and only 11\% rated as 3 on a 0-5 scale; no response was rated lower than 3. These findings are consistent with a prior study in which ChatGPT responses were reported as having high quality~\cite{alyssa2023assessing} to real-world questions asked by dementia caregivers. 

We found no interaction effect of system and task prompts or main effect of system prompts on response length. However, we did find a statistically significant effect of task prompts on response length, with TP3, which instructed GPT-4 to generate a response ``as long as possible with as much detail as possible,`` producing the lengthiest responses. This was expected, because we designed the 3 task prompts to instruct GPT-4 to generate responses with varying levels of detail.  Also as expected, GPT-4 followed our specified 3-component structure in TP2 to present educational information, tangible actions, and referrals sequentially. (While responses without TP 2 also produced content with these 3 components, they were not as structured and less explicit.) This finding suggests that, consistent with previous research~\cite{singhal2022large}, the CoT in our initialization prompt was useful in guiding GPT-4 to produce structured answers by incorporating a structured reasoning process and specific references in some cases.

Our statistical analysis found no interaction or main effect of system or task prompts on response quality. Our clinicians' qualitative feedback provided insight to understand these findings. They indicated that system prompts with different roles might not have had much, if any, impact on response content. In the case where a recommendation for  ``a full cognitive assessment`` was generated, it was not generated with the role of a neuropsychologist. Instead, it was recommended in the response associated with the role of \textit{a helpful assistant}. This finding differs with Zheng’s~\cite{zheng2023helpful}, in which adding roles in prompts consistently improved model performance. Unlike general social roles such as ``farmer,`` ``police,`` or ``doctor`` in Zheng’s study, the roles in our system prompts were specific to the dementia care domain (i.e., ``neuropsychologist`` and ``social worker``). These specific roles were likely absent in the prompts or data during the training of GPT-4; thus, GPT-4 may be incapable of differentiating these specialties. Also, Zheng's study used only the prompt ``You are a [role]``, whereas our system prompts included roles with specific descriptions. Another plausible explanation for the absence of any role effect in our results might be attributable to the specificity of our tasks. Dementia is prevalent among older adults, but it remains a niche field compared with more widely studied natural language processing tasks such as name entity recognition in training a general LLM. Therefore, it is conceivable that GPT-4 might not have been exposed to sufficient dementia-specific circumstances to learn how to react to them, particularly in handling open-ended questions and generating responses to caregivers.

Also, our clinicians noted that TP3, the task prompt that instructed LLMs to produce responses “as long as possible with as much detail as possible”, might not necessarily have produced better quality responses, clinically, as some of the detailed information may not be necessary for caregivers. This suggests that if an LLM complies with a prompt to generate responses ``as long as possible with as much detail as possible``, it may do so at the cost of adding unnecessary details. A growing concern in the LLM literature is that LLMs tend to fabricate or hallucinate in their responses~\cite{ahmad2023creating}. In our study, the clinicians found all responses being factual, suggesting no fabrication or hallucination in GPT-4 responses. However, even when the responses, including the excessive information, were medically accurate, they could still be unnecessary or not useful. Too much unnecessary information may distract and overwhelm caregivers, who are known to be stressed and often pressed for time. Future research should systematically explore the right balance between generating too little and too much information, and identifying the best prompts that lead to the right amount and types of information for caregivers with varying preferences and situations.

This exploratory study has limitations, which call for future research.
\begin{enumerate}
    \item We used a small, convenience sample—only 3 posts, 1 per area (memory loss and confusion; aggression; and driving) to generate responses. With such a small sample, statistical analysis was not as conclusive as it would have been with a large sample. Future research would benefit from using a large sample of posts that represent more types of HIWs that caregivers might have.
    \item We included only 1 clinician in neuropsychology and 1 in social work to evaluate the quality of GPT-4 responses. Future research would benefit from involving more clinicians in additional specialized areas to explore potential differences across clinical specialties (e.g., nursing, neurology, and pharmacy).
    \item By having experienced clinicians assess the quality of GPT-4 responses, we were able to ensure the quality of the responses from a clinical perspective, thus ensuring that the responses were safe and did not contain harmful, misleading information. Such an assessment, however, cannot indicate whether or how caregivers might actually find responses useful for overcoming the challenges they experience. Future research should include caregivers’ assessments of AI-generated responses in terms of usefulness.
    \item Our rating scale\cite{alyssa2023assessing} was initially developed to assess the quality of AI-generated responses to different posts with the same prompt, SP1. It was not designed to compare the quality of AI-generated responses to the same posts with different prompts. Given that AI-generated responses received high scores across the board, using this rating scale might have amplified a ceiling effect, obscuring potentially subtle differences and nuances. It will be useful in future research to develop a more applicable rating scale to ensure comprehensive comparisons of responses with different prompts.

\end{enumerate}

\section{Conclusions}
In this exploratory study, we examined the impact of different combinations of prompts on GPT-4 generated responses to dementia caregivers’ questions. We developed a prompt template with 12 combinations of prompts: system prompts (4 roles), one initialization prompt, and task prompts (requesting 3 levels of details), and we evaluated their performance in generating responses to complex dementia caregiving queries. We found that instructing GPT-4 to generate detailed responses led to statistically significantly longer responses, but at times the details contained in those lengthy responses did \textit{not} seem necessary from a clinical perspective. System prompts with different professional roles did not generate noticeable differences in response length or quality. However, our study was limited by its small sample size, with only two types of clinical expertise represented for evaluation, and lacked caregiver involvement. Future research will benefit from a systematic investigation of the effects of different prompts using larger samples and comprehensive evaluation metrics that include both clinicians’ and caregivers’ assessments.

\section*{Acknowledgment}
This work was in part supported by the National Institute on Aging of the National Institutes of Health under Award Number R56AG075770. The content is solely the responsibility of the authors and does not necessarily represent the official views of the National Institutes of Health. Editorial support with manuscript development was provided by Dr. John Bellquist at the Cain Center for Nursing Research at The University of Texas at Austin School of Nursing.

\section*{APPENDIX}
\label{appendixA}
\subsection*{Appendix A: The full prompt template} 

\begin{itemize}
\item System Prompts
\begin{itemize}
\item System prompt 1 (SP1). \textit{You are a helpful assistant.}

\item System prompt 2 (SP2). \textit{You are an experienced clinician specialized in dementia care.}

\item System prompt 3 (SP3). \textit{You are a board-certified clinical neuropsychologist in the Department of Neurology at a large Medical School. You have been the Director of clinics serving Alzheimer's disease and related dementias (ADRD) patients and their caregivers for over 15 years. Currently, you are the Director of a Comprehensive Memory Center, which is a collaborative, interprofessional team-based dementia clinic. }

\item System prompt 4 (SP4). \textit{You are a LCSW-S and Assistant Director of Dementia Care Transformation in a Comprehensive Memory Center, which is a collaborative, interprofessional team-based dementia clinic. As a licensed clinical social worker for 15 years, you have implemented evidence based ADRD interventions and supported family caregivers. You have helped design patient-centered approaches to dementia care and conducted patient and caregiver focus groups and have facilitated 500+ counseling sessions.}
\end{itemize}
\item Initialization prompt: \textit{I will give you tasks that focus on providing tailored, evidence-based care strategies to help informal dementia caregivers manage the daily care for a family member living with dementia. Always answer as helpful as possible, while being safe and avoiding giving harmful advice or information. Let’s think step by step, referring to authoritative sources as needed. Do you understand? }

\item Task prompt:
\begin{itemize}
\item Task prompt 1 (TP1). \textit{Generate a response to the text in brackets: [post].}

\item Task prompt 2 (TP2). \textit{Generate a response to the text in brackets: [post]. In generating your response, follow the following 3-component structure: first, educational information; second, tangible actions the caregiver can take; and finally, referrals that suggest helpful resources for the caregiver. }

\item Task prompt 3 (TP3). \textit{Generate a response to the text in brackets: [post]. In generating your response, follow the following 3-component structure: first, educational information; second, tangible actions the caregiver can take; and finally, referrals that suggest helpful resources for the caregiver. Try to generate the response as long as possible with as much detail as possible to cover all relevant aspects with all available specifics. Do not limit the length of the response to any word limit. }
\end{itemize}
\end{itemize}

\bibliographystyle{IEEEtran}
\bibliography{citations}

\end{document}